\documentclass {article}  		 		% Specifies the document style.
\usepackage{index, algorithm}				% Specifies the index style.	
\usepackage{hyperref}
\usepackage{graphicx}

\begin{document}           				% End of preamble and beginning of text.

\title{Improving the Search by Encoding Multiple Solutions in a Chromosome}	% Title \begin{center}
\author{Mihai Oltean\\
\\
Department of Computer Science, \\
Faculty of Mathematics and Computer Science, \\
Babe\c s Bolyai University, \\
Kogalniceanu 1, 3400 Cluj-Napoca, Romania, \\
mihai.oltean@gmail.com}

%%%%%%%%%%%%%%%%%%%%%%%%%%%%%%%%%%%%%%%%%%%%%%%%%%%%%%%%%
%							%
%    HERE STARTS THE MAIN CONTENTS OF THE CHAPTER	%
%							%
%%%%%%%%%%%%%%%%%%%%%%%%%%%%%%%%%%%%%%%%%%%%%%%%%%%%%%%%%
\date{}

\maketitle

\begin{abstract}

We investigate the possibility of encoding multiple solutions of a problem in a single chromosome. The best solution encoded in an individual will represent (will provide the fitness of) that individual. In order to obtain some benefits the chromosome decoding process must have the same complexity as in the case of a single solution in a chromosome. Three Genetic Programming techniques are analyzed for this purpose: Multi Expression Programming, Linear Genetic Programming and Infix Form Genetic Programming. Numerical experiments show that encoding multiple solutions in a chromosome greatly improves the search process.

\end{abstract}

\section{Introduction}

Evolutionary Algorithms (EAs) \cite{15:goldberg1,15:holland1} are powerful tools used for solving difficult real-world problems.

This paper describes a new paradigm called \textit{Multi Solution Programming} (MSP) that may be used for improving the search performed by the Evolutionary Algorithms. The main idea is to encode multiple solutions (more than one) in a single chromosome. The best solution encoded in a chromosome will represent (will provide the fitness of) that individual.

This special kind of encoding is useful when the complexity of the decoding process is similar to the complexity of the decoding process of chromosomes encoding a single solution of the problem being solved.

Note that the Multi Solution Programming is not a particular technique, but a paradigm intended to be used in conjunction with an Evolutionary Algorithm. MSP refers to a new way of encoding solutions in a chromosome.

GP techniques are very suitable for the MSP paradigm because the trees offer an implicit multiple solutions representation: each sub-tree may be considered as a potential solution of the problem. Three Genetic Programming (GP) \cite{15:koza1,15:koza2} variants are tested with the proposed model: Multi Expression Programming (MEP) \cite{15:oltean1,15:oltean2}, Linear Genetic Programming (LGP) \cite{15:brameier1,15:brameier2,15:brameier3,15:nordin1} and Infix Form Genetic Programming (IFGP) \cite{15:oltean3}.

Multi Expression Programming uses a chromosome encoding similar to the way in which \textit{C} or \textit{Pascal} compilers translate mathematical expressions in machine code. Note that MEP was originally designed \cite{15:oltean1} to encode multiple solutions in a chromosome. A MEP variant that encodes a single solution in a chromosome is proposed and tested in this paper. 

A Linear Genetic Programming chromosome is a sequence of \textbf{\textit{C}} language instructions. Each instruction has a destination variable and several source variables. One of the variables is usually chosen to provide the output of the program. A LGP variant encoding multiple solutions in a chromosome is proposed in this paper.

Infix Form Genetic Programming chromosomes are strings of integer encoding mathematical expressions in infix form. Each IFGP chromosome encodes multiple solutions of the problem being solved \cite{15:oltean3}. A IFGP variant encoding a single solution in a chromosome is proposed and tested in this paper.

All the solutions represented in a MSP individual should be decoded by traversing the chromosome only once. Partial 
results are stored by using Dynamic Programming. 

Several numerical experiments with the considered GP techniques are performed by using 4 
symbolic regression problems. For each test problem the relationships between the success 
rate and the population size and the code length are analyzed. 

Results show that Multi Solutions Programming significantly improves the evolutionary search. For all considered test 
problems the evolutionary techniques encoding multiple solutions in a chromosome are significantly better than the similar techniques encoding a single solution in a chromosome.

The paper is organized as follows. Motivation for this research is presented in section \ref{sec:15-1}. Test problems used to asses the performance of the considered algorithms are given in section \ref{sec:15-2}. Multi Expression Programming and its counterpart Single Expression Programming are described in section \ref{sec:15-3}. Linear Genetic Programming and Multi-Solution Linear Genetic Programming are described in section \ref{sec:15-4}. Infix Form Genetic Programming and Single-Solution Infix Form Genetic Programming are described in section \ref{sec:15-5}. Conclusions and further work directions are given in section \ref{sec:15-6}.

\section{Multiple Solution Programming} \label{sec:15-1}

This section tries to answer two fundamental questions:

\begin{itemize}
\item{Why encoding multiple solutions in the same chromosome?}
\item{How to efficiently encode multiple solutions in the same chromosome in order to obtain some benefits?}
\end{itemize}

The answer for the first question is motivated by the No Free Lunch Theorems for Search 
\cite{15:wolpert1}. There is neither practical nor theoretical evidence that one of the 
solutions encoded in a chromosome is better than the others. More than that, 
Wolpert and McReady \cite{15:wolpert1} proved that we cannot use the search algorithm's 
behavior so far for a particular test function to predict its future 
behavior on that function.

The second question is more difficult than the first one. There is no general prescription on how to encode multiple solutions in the same chromosome. In most cases this involves creativity and imagination. However, some general suggestions can be given.

\begin{itemize}

\item[{\it (i)}]{In order to obtain some benefits from encoding more than one solution in a chromosome we have to spend a similar effort (computer time, memory etc) as in the case of encoding a single solution in a chromosome. For instance, if we have 10 solutions encoded in chromosome and the time needed to extract and decode these solutions is 10 times larger than the time needed to extract and process one solution we got nothing. In this case we can not talk about an useful encoding.}

\item[{\it (ii)}]{We have to be careful when we want to encode a multiple solutions in a variable length chromosome (for instance in a standard GP chromosome), because this kind of chromosome will tend to increase its size in order to encode more and more solutions. And this could lead to bloat \cite{15:banzhaf1,15:luke1}.}

\item[{\it (iii)}]{Usually encoding multiple solutions in a chromosome might require storing the partial results. Sometimes this can be achieved by using the Dynamic Programming \cite{15:bellman1} technique. This kind of model is employed by the techniques described in this paper.}

\end{itemize}

\section{Test Problems and Metric of Performance}\label{sec:15-2}

For assessing the performance of the considered algorithms we use several symbolic regression problems \cite{15:koza1}. The aim of symbolic regression is to find a mathematical expression that satisfies a set of fitness cases. 

The test problems used in the numerical experiments are:\\

$F_{1}(x)=x^{4}-x^{3}+x^{2}-x$,\\

$F_{2}(x)=x^{4}+x^{3}+x^{2}+x$,\\

$F_{3}(x)=x^{4}+2*x^{3}+3*x^{2}+4*x$,\\

$F_{4}(x)=x^{6}-2x^{4}+x^{2}$.\\

Test problem $f_{2}$ is also known as \textit{quartic polynomial} and $f_{4}$ is known as \textit{sextic polynomial} \cite{15:koza1,15:koza2}.

For each function 20 fitness cases have been randomly generated with a 
uniform distribution over the [0, 10] interval.

For each test problem the relationships between the success rate, the population size and the code length are analyzed. The success rate is computed as:

\begin{equation}
Success\,rate = 
\frac{The\,number\,of\,successful\,runs}{The\,total\,number\,of\,runs}.
\end{equation}

Roughly speaking the quality of a GP chromosome is the average distance between the expected output and the obtained output \cite{15:koza1}. A run is considered successful if the fitness of the best individual in the last generation is less than 0.01.

\section{Multi Expression Programming}\label{sec:15-3}

In this section, \textit{Multi Expression Programming} (MEP) \cite{15:oltean1} is briefly described. MEP source code can be downloaded from \url{https://mepx.org}

\subsection{MEP Representation}\label{sec:15-3-1}

MEP genes are represented by substrings of a variable length. The number of 
genes per chromosome is constant. This number defines the length of the 
chromosome. Each gene encodes a terminal or a function symbol. A gene 
encoding a function includes pointers towards the function arguments. 
Function arguments always have indices of lower values than the position of 
that function in the chromosome.

This representation is similar to the way in which \textbf{\textit{C}} and 
\textbf{\textit{Pascal}} compilers translate mathematical expressions into 
machine code \cite{15:aho1}.

MEP representation ensures that no cycle arises while the 
chromosome is decoded (phenotypically transcripted). According to
this representation scheme the first symbol of the chromosome must be a 
terminal symbol. In this way only syntactically correct programs (MEP 
individuals) are obtained.\\

\textbf{Example}\\

A representation where the numbers on the left positions stand for gene 
labels is employed here. Labels do not belong to the chromosome, they being 
provided only for explanation purposes.

For this example we use the set of functions $F$ = {\{}+, *{\}}, and the set of 
terminals $T$ = {\{}$a$, $b$, $c$, $d${\}}. An example of chromosome using the sets $F$ and 
$T$ is given below:\\

1: $a$

2: $b$

3: + 1, 2

4: $c$

5: $d$

6: + 4, 5

7: * 3, 6

\subsection{Decoding MEP Chromosomes and Fitness Assignment Process}\label{sec:15-3-2}

In this section is described the way in which MEP individuals are 
translated into computer programs and the way in which the fitness of these 
programs is computed.

This translation is achieved by reading the chromosome top-down. A terminal 
symbol specifies a simple expression. A function symbol specifies a complex 
expression obtained by connecting the operands specified by the argument 
positions with the current function symbol.

For instance, genes 1, 2, 4 and 5 in the previous example encode simple 
expressions formed by a single terminal symbol. These expressions are:\\

$E_{1}=a$,

$E_{2}=b$,

$E_{4}=c$,

$E_{5}=d$,\\

Gene 3 indicates the operation + on the operands located at positions 1 and 
2 of the chromosome. Therefore gene 3 encodes the expression: \\

$E_{3}=a+b$. \\

Gene 6 indicates the operation + on the operands located at positions 4 and 
5. Therefore gene 6 encodes the expression:\\

$E_{6}=c+d$.\\

Gene 7 indicates the operation * on the operands located at position 3 and 
6. Therefore gene 7 encodes the expression:\\

$E_{7}$ = ($a+b)$ * ($c+d)$.\\

$E_{7}$ is the expression encoded by the whole chromosome.

Each MEP chromosome is allowed to encode a number of expressions 
equal to the chromosome length. Each of these expressions is considered as 
being a potential solution of the problem. 

The value of these expressions may be computed by reading the chromosome top 
down. Partial results are computed by dynamic programming and are stored in 
a conventional manner.

Usually the chromosome fitness is defined as the fitness of the best 
expression encoded by that chromosome.

For instance, if we want to solve symbolic regression problems the fitness 
of each sub-expression $E_{i}$ may be computed using the formula:

\begin{equation}
\label{eq1}
f(E_i ) = \sum\limits_{k = 1}^n {\left| {o_{k,i} - w_k } \right|} ,
\end{equation}

\noindent
where $o_{k,i}$ is the obtained result by the expression $E_{i}$ for the 
fitness case $k$ and w$_{k}$ is the targeted result for the fitness case $k$. In 
this case the fitness needs to be minimized.

The fitness of an individual is set to be equal to the lowest fitness of the 
expressions encoded in chromosome:

\begin{equation}
\label{eq2}
f(C) = \mathop {\min }\limits_i f(E_i ).
\end{equation}

When we have to deal with other problems we compute the fitness of each 
sub-expression encoded in the MEP chromosome and the fitness of the entire 
individual is given by the fitness of the best expression encoded in that 
chromosome.

\subsection{Search Operators}

Search operators used within MEP algorithm are crossover and mutation. 
Considered search operators preserve the chromosome structure. All offspring 
are syntactically correct expressions. 

\subsubsection{Crossover}

By crossover two parents are selected and are recombined. For instance, 
within the uniform recombination the offspring genes are taken randomly from 
one parent or another.\\

\textbf{Example}\\

Let us consider the two parents $C_{1}$ and $C_{2}$ given in Table \ref{table:15-1}. The two 
offspring $O_{1}$ and $O_{2}$ are obtained by uniform recombination as 
shown in Table \ref{table:15-1}.

\begin{table}[htbp]
\caption{MEP uniform recombination.}
\label{table:15-1}
\begin{center}
\begin{tabular}
{p{55pt}p{49pt}p{49pt}p{49pt}}
\hline
\multicolumn{2}{p{104pt}}{\textit{Parents}} & 
\multicolumn{2}{p{99pt}}{\textit{Offspring}}  \\
%\hline
$C_{1}$& 
$C_{2}$& 
$O_{1}$& 
$O_{2}$ \\
\hline
1: \textbf{\textit{b}} \par 2: \textbf{* 1, 1} \par 3: \textbf{+ 2, 1} \par 4: \textbf{\textit{a}} \par 5: \textbf{* 3, 2} \par 6: \textbf{\textit{a}} \par 7: \textbf{- 1, 4}& 
1: $a$ \par 2: $b$ \par 3: + 1, 2 \par 4: $c$ \par 5: $d$ \par 6: + 4, 5 \par 7: * 3, 6& 
1: $a$ \par 2: \textbf{* 1, 1} \par 3: \textbf{+ 2, 1} \par 4: $c$ \par 5: \textbf{* 3, 2} \par 6: + 4, 5 \par 7: \textbf{- 1, 4}& 
1: \textbf{\textit{b}} \par 2: $b$ \par 3: + 1, 2 \par 4: \textbf{\textit{a}} \par 5: $d$ \par 6: \textbf{\textit{a}} \par 7: * 3, 6 \\
\hline
\end{tabular}
\end{center}
\end{table}

\subsubsection{Mutation}

Each symbol (terminal, function of function pointer) in the chromosome may 
be target of mutation operator. By mutation some symbols in the chromosome 
are changed. To preserve the consistency of the chromosome its first gene 
must encode a terminal symbol.\\

\textbf{Example}\\

Consider the chromosome $C$ given in Table \ref{table:15-2}. If the boldfaced symbols are selected 
for mutation an offspring $O$ is obtained as shown in Table \ref{table:15-2}.

\begin{table}[htbp]
\caption{MEP mutation.}
\label{table:15-2}
\begin{center}
\begin{tabular}
{p{81pt}p{67pt}}
\hline
$C$& 
$O$ \\
\hline
1: $a$ \par 2: * 1, 1 \par 3: \textbf{\textit{b}} \par 4: * 2, 2 \par 5: $b$ \par 6: +\textbf{ 3}, 5 \par 7: $a$& 
1: $a$ \par 2: * 1, 1 \par 3: \textbf{+ 1, 2} \par 4: * 2, 2 \par 5: $b$ \par 6: \textbf{+ 1}, 5 \par 7: $a$ \\
\hline
\end{tabular}
\end{center}
\end{table}

\subsection{MEP Algorithm}

In this paper we use a steady-state \cite{15:syswerda1} as underlying mechanism for Multi 
Expression Programming. The algorithm starts by creating a random population 
of individuals. The following steps are repeated until a stop condition is 
reached. Two parents are selected using a selection procedure. The parents 
are recombined in order to obtain two offspring. The offspring are 
considered for mutation. The best offspring replaces the worst individual in 
the current population if the offspring is better than the worst individual. 

The algorithm returns as its answer the best expression evolved along a 
fixed number of generations.

\subsection{Single Expression Programming}

The MEP variant encoding a single solution in a chromosome is called Single Expression Programming (SEP). The expression encoded in a SEP chromosome is given by its last gene.\\

\textbf{Example}\\

Consider again the chromosome given in section \ref{sec:15-3-1}.\\

1: $a$

2: $b$

3: + 1, 2

4: $c$

5: $d$

6: + 4, 5

7: * 3, 6\\

The SEP expression encoded by this chromosome is:\\

$E=(a+b)*(c+d)$.

\subsection{Numerical Experiments with MEP and SEP}

Several numerical experiments with Multi Expression Programming and Single Expression Programming are performed in this section using the test problems described in section \ref{sec:15-2}.

The general parameters of the MEP and SEP algorithms are given in Table \ref{table:15-3}. The same 
settings are used for Multi Expression Programming and for Single Expression Programming.

\begin{table}[htbp]
\begin{center}
\caption{Parameters of the MEP and SEP algorithms for solving symbolic regression problems.}
\label{table:15-3}
\begin{tabular}
{p{130pt}p{130pt}}
\hline
\textbf{Parameter}& 
\textbf{Value} \\
\hline
Number of generations& 
51 \\
%\hline
Crossover probability& 
0.9 \\
%\hline
Mutations& 
2 / chromosome \\
%\hline
Function set& 
$F$ = \{+, -, *, /, \} \\
%\hline
Terminal set& 
Problem inputs\\
%\hline
Selection& 
Binary Tournament \\
\hline
\end{tabular}
\end{center}
\end{table}

For all problems the relationship between the success rate and the 
chromosome length and the population size is analyzed. The success rate is 
computed as the number of successful runs over the total number of runs (see section \ref{sec:15-2}). \\

\textbf{Experiment 1}\\

\begin{figure}[htbp]
\centerline{\includegraphics[width = \textwidth]{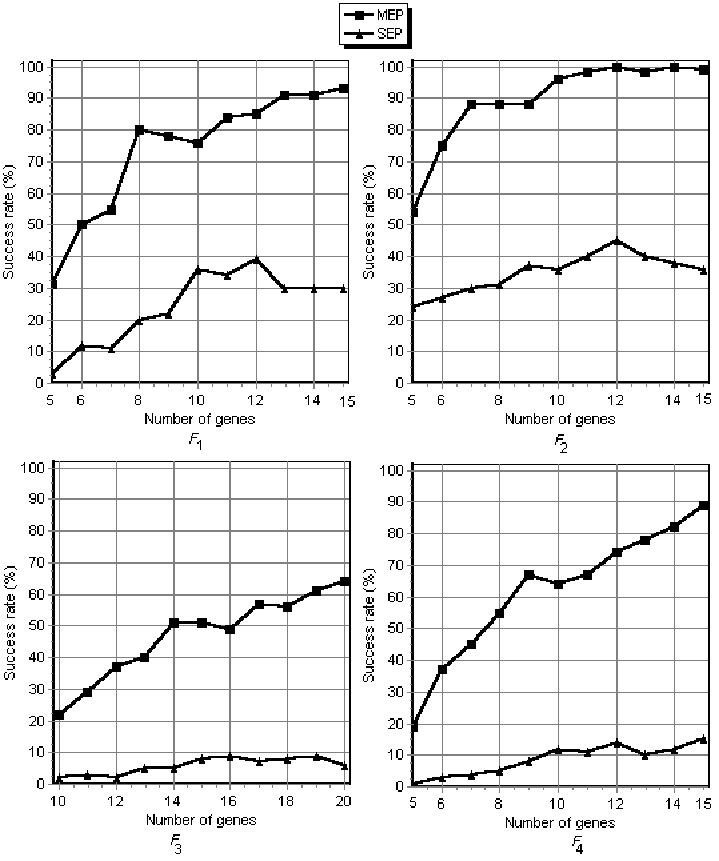}}
\caption{The relationship between the success rate and the number 
of instructions in a chromosome. Results are averaged over 100 runs.}
\label{fig:15-1}
\end{figure}

In this experiment the relationship between the success rate and the 
chromosome length is analyzed. The population size was 
set to 50 individuals. Other parameters of the MEP and SEP algorithms are given in 
Table \ref{table:15-3}. Results are depicted in Figure \ref{fig:15-1}.

Figure \ref{fig:15-1} shows that Multi Expression Programming significantly outperforms 
Single Expression Programming  for all the considered test problems and for all the 
considered parameter setting. More than that, large chromosomes are better 
for MEP than short chromosomes. This is due to the multi-solution 
ability: increasing the chromosome length leads to more solutions encoded in 
the same individual. The easiest problem is $f_{2}$. MEP success rate for this problem is over 90{\%} when the number of instructions in a chromosome is larger than 10. 
The most difficult problem is $f_{3}$. For this problem and with the 
parameters given in Table \ref{table:15-3}, the success rate of the MEP algorithm never 
increases over 70{\%}. However, these results are very good compared to 
those obtained by SEP (the success rate never increases over 10{\%} for the test problem $f_{3}$).\\

\textbf{Experiment 2}\\

\begin{figure}[htbp]
\centerline{\includegraphics[width = \textwidth]{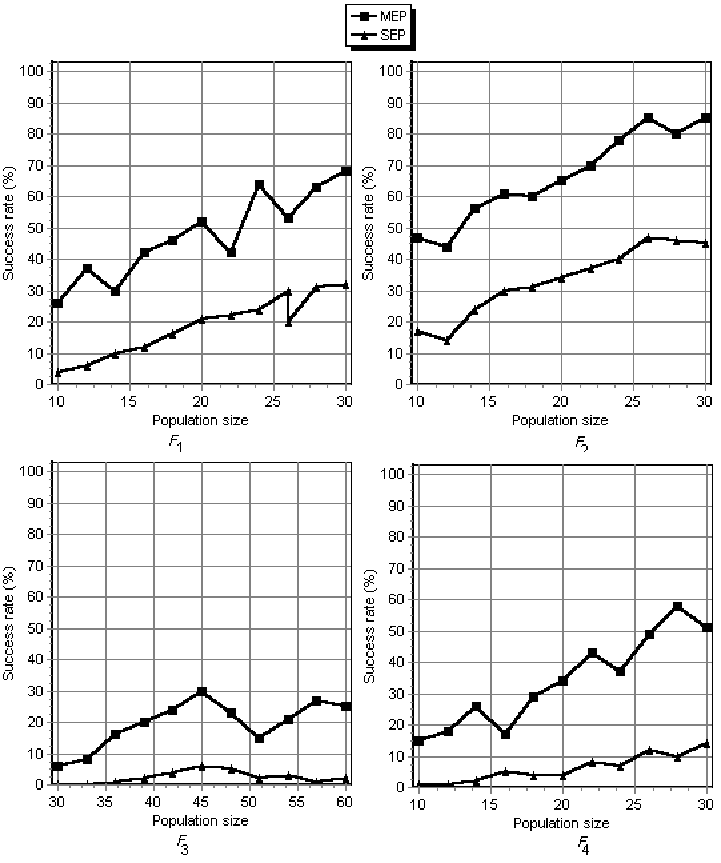}}
\caption{The relationship between the success rate and the population size. Results are averaged over 100 runs.}
\label{fig:15-2}
\end{figure}

In this experiment the relationship between the success rate and the 
population size is analyzed. The number of instructions 
in a MEP or SEP chromosome was set to 10. Other parameters for the MEP and SEP are given in Table \ref{table:15-3}. Results are depicted in Figure \ref{fig:15-2}.

Figure \ref{fig:15-2} shows that MEP performs better than SEP. Problem $f_{2}$ is the easiest one and problem $f_{3}$ is the most difficult one.

\section{Linear Genetic Programming}
\label{sec:15-4}

\textit{Linear Genetic Programming} (LGP) \cite{15:brameier1,15:nordin1} uses a specific linear representation of computer programs. Instead of the tree-based GP expressions \cite{15:koza1} of a functional programming 
language (like \textbf{\textit{LISP}}), programs of an imperative language 
(like \textbf{\textit{C}}) are evolved.

\subsection{LGP Representation}

A LGP individual is represented by a variable-length sequence of simple 
\textbf{\textit{C}} language instructions. Instructions operate on one or 
two indexed variables (registers) $r$ or on constants $c$ from predefined sets. 
The result is assigned to a destination register, e.g. $r_{i}=r_{j}$ * 
$c$.

An example of the LGP program is the following one:\\

\textbf{void} LGP(\textbf{double} $r$[8])

{\{}

\hspace{.5cm}$r[0] = r[5] + 73$;

\hspace{.5cm}$r[7] = r[3] - 59$;

\hspace{.5cm}$r[2] = r[5] + r[4]$;

\hspace{.5cm}$r[6] = r[7] * 25$;

\hspace{.5cm}$r[1] = r[4] - 4$;

\hspace{.5cm}$r[7] = r[6] * 2$;

{\}}

\subsection{Decoding LGP Individuals}

A linear genetic program can be turned into a functional representation by 
successive replacements of variables starting with the last effective 
instruction \cite{15:brameier1}.

Usually one of the variables ($r$[0]) is chosen as the output of the program. 
This choice is made at the beginning of the program and is not changed 
during the search process. In what follows we will denote this LGP variant 
as Single-Solution Linear Genetic Programming (SS-LGP).

\subsection{Genetic Operators}

The variation operators used in conjunction with Linear Genetic Programming 
are crossover and mutation. Standard LGP crossover works by exchanging 
continuous sequences of instructions between parents \cite{15:brameier1}.

Two types of standard LGP mutations are usually used: micro mutation and 
macro mutation. By micro mutation an operand or an operator of an 
instruction is changed \cite{15:brameier1}.

Macro mutation inserts or deletes a random instruction \cite{15:brameier1}.

Since we are interested more in multi-solutions paradigm rather than in 
variable length chromosomes we will use fixed length chromosomes in all 
experiments performed in this paper. Genetic operators used in numerical 
experiments are uniform crossover and micro mutation.

\subsubsection{LGP uniform crossover}

LGP uniform crossover works between instructions. The offspring's genes 
(instructions) are taken with a 50{\%} probability from the parents.\\

\textbf{Example}\\

Let us consider the two parents $C_{1}$ and $C_{2}$ given in Table \ref{tabel:15-4}. The two 
offspring $O_{1}$ and $O_{2}$ are obtained by uniform recombination as shown in Table \ref{tabel:15-4}.

\begin{table}[htbp]
\begin{center}
\caption{LGP uniform recombination.}
\label{tabel:15-4}
\begin{tabular}
{p{80pt}p{90pt}p{90pt}p{90pt}}
\hline
\multicolumn{2}{p{170pt}}{Parents} & 
\multicolumn{2}{p{190pt}}{Offspring}  \\
\hline
$C_{1}$& 
$C_{2}$& 
$O_{1}$& 
$O_{2}$\\
\hline
$r$[5]=$r$[3]*$r$[2]; \par $r$[3]=$r$[1]+6; \par $r$[0]=$r$[4]*$r$[7]; \par $r$[5]=$r$[4]-$r$[1]; \par $r$[1]=$r$[6]*7; \par $r$[0]=$r$[0]+$r$[4]; \par $r$[2]=$r$[3]/$r$[4];&
\textbf{\textit{r}}\textbf{[2]=}\textbf{\textit{r}}\textbf{[0]+r[3];} \par \textbf{\textit{r}}\textbf{[1]=}\textbf{\textit{r}}\textbf{[2]*r[6];} \par \textbf{\textit{r}}\textbf{[4]=}\textbf{\textit{r}}\textbf{[6]-4;} \par \textbf{\textit{r}}\textbf{[6]=}\textbf{\textit{r}}\textbf{[5]/r[2];} \par \textbf{\textit{r}}\textbf{[2]=}\textbf{\textit{r}}\textbf{[1]+7;} \par \textbf{\textit{r}}\textbf{[1]=}\textbf{\textit{r}}\textbf{[2]+r[4];} \par \textbf{\textit{r}}\textbf{[0]=}\textbf{\textit{r}}\textbf{[4]*3;}&
$r$[5]=$r$[3]*$r$[2]; \par \textbf{\textit{r}}\textbf{[1]=}\textbf{\textit{r}}\textbf{[2]*r[6];} \par $r$[0]=$r$[4]*$r$[7]; \par $r$[5]=$r$[4]-$r$[1]; \par \textbf{\textit{r}}\textbf{[2]=}\textbf{\textit{r}}\textbf{[1]+7;} \par \textbf{\textit{r}}\textbf{[1]=}\textbf{\textit{r}}\textbf{[2]+r[4];} \par \textbf{\textit{r}}\textbf{[0]=}\textbf{\textit{r}}\textbf{[4]*3;}& 
\textbf{\textit{r}}\textbf{[2]=}\textbf{\textit{r}}\textbf{[0]+r[3];} \par $r$[3]=$r$[1]+6; \par \textbf{\textit{r}}\textbf{[4]=}\textbf{\textit{r}}\textbf{[6]-4;} \par \textbf{\textit{r}}\textbf{[6]=}\textbf{\textit{r}}\textbf{[5]/r[2];} \par $r$[1]=$r$[6]*7; \par $r$[0]=$r$[0]+$r$[4]; \par $r$[2]=$r$[3]/$r$[4]; \par \\
\hline
\end{tabular}
\end{center}
\end{table}

\subsubsection{LGP mutation}

LGP mutation works inside of a LGP instruction. By mutation each operand 
(source or destination) or operator is affected with a fixed mutation 
probability.\\

\textbf{Example}\\

Consider an individual $C$ which is affected by mutation. An offspring $O$ is 
obtained as shown in Table \ref{tabel:15-5} (modified variables are written in boldface):

\begin{table}[htbp]
\begin{center}
\caption{LGP mutation.}
\label{tabel:15-5}
\begin{tabular}
{p{95pt}p{94pt}}
\hline
$C$& 
$O$ \\
\hline
$r$[5] = $r$[3] * $r$[2]; \par $r$[3] = $r$[1] + 6; \par $r$[0] = $r$[4] * $r$[7]; \par $r$[5] = $r$[4] - $r$[1]; \par $r$[1] = $r$[6] * 7; \par $r$[0] = $r$[0] + $r$[4]; \par $r$[2] = $r$[3] / $r$[4];& 
$r$[5] = $r$[3] * $r$[2]; \par $r$[3] = \textbf{\textit{r}}\textbf{[6]} + \textbf{\textit{r}}\textbf{[0]}; \par $r$[0] = $r$[4] \textbf{+} $r$[7]; \par \textbf{\textit{r}}\textbf{[4]} = $r$[4] - $r$[1]; \par $r$[1] = $r$[6] * \textbf{2}; \par $r$[0] = $r$[0] + $r$[4]; \par \textbf{\textit{r}}\textbf{[0]} = $r$[3] / $r$[4]; \\
\hline
\end{tabular}
\end{center}
\end{table}

\subsection{LGP Algorithm}

LGP uses a modified steady-state algorithm \cite{15:brameier1}. Initial population is randomly 
generated. The following steps are repeated until a termination criterion is 
reached: Four individuals are randomly selected from the current population. 
The best two of them are considered the winners of the tournament and they 
will act as parents. The parents are recombined and the offspring are 
mutated and then replace the losers of the tournament.

\subsection{Multi Solution Linear Genetic Programming}

The LGP structure is enriched as follows:

\begin{itemize}
\item[{\it (i)}] {We allow as each destination variable to represent the output of the 
program. In the standard LGP only one variable is chosen to provide the 
output. }

\item[{\it (ii)}] {We check for the program output after each instruction in chromosome. This is again different from the standard LGP where the output was checked after the execution of all instructions in a chromosome.}

\end{itemize}

After each instruction, the value stored in the destination variable is 
considered as a potential solution of the problem. The best value stored in 
one of the destination variables is considered for fitness assignment 
purposes. \\

\textbf{Example}\\

Consider the chromosome $C$ given below:\\

\textsf{\textbf{void}}\textsf{ LGP(}\textsf{\textbf{double}}\textsf{ 
}\textsf{\textit{r}}\textsf{[8])}

\textsf{{\{}}

\hspace{.5cm}$r[5] = r[3] * r[2]$;

\hspace{.5cm}$r[3] = r[1] + 6$;

\hspace{.5cm}$r[0] = r[4] * r[7];$

\hspace{.5cm}$r[6] = r[4] - r[1];$

\hspace{.5cm}$r[1] = r[6] * 7;$

\hspace{.5cm}$r[2] = r[3] / r[4];$

{\}}\\

Instead of encoding the output of the problem in a single variable (as in 
SS-LGP) we allow that each of the destination variables ($r$[5], $r$[3], $r$[0], 
$r$[6], $r$[1] or $r$[2]) to store the program output. The best output stored in 
these variables will provide the fitness of the chromosome.

For instance, if we want to solve symbolic regression problems, the fitness 
of each destination variable $r$[$i$] may be computed using the formula:

\[
f(r[i]) = \sum\limits_{k = 1}^n {\left| {o_{k,i} - w_k } \right|} ,
\]

\noindent
where $o_{k,i}$ is the result obtained in variable $r$[$i$] for the fitness case 
$k$, $w_{k}$ is the targeted result for the fitness case $k$ and $n$ is the number of 
fitness cases. For this problem the fitness needs to be minimized.

The fitness of an individual is set to be equal to the lowest fitness of the 
destination variables encoded in the chromosome:

\[
f(C) = \mathop {\min }\limits_i f(r[i]).
\]

Thus, we have a Multi-Solution LGP program at two levels: first level is given 
by the possibility that each variable to represent the output of the program 
and the second level is given by the possibility of checking for the output 
at each instruction in the chromosome.

\subsection{Numerical Experiments with LGP and MS-LGP}

In this section several experiments with SS-LGP and MS-LGP are performed. 
For this purpose we use several well-known symbolic regression problems described in section \ref{sec:15-2}.

The general parameters of the LGP algorithms are given in Table \ref{tabel:15-6}. The same 
settings are used for Multi-Solution LGP and for Single-Solution LGP.

\begin{table}[htbp]
\begin{center}
\caption{Parameters of the LGP and MS-LGP algorithms for solving symbolic regression problems.}
\label{tabel:15-6}
\begin{tabular}
{p{120pt}p{210pt}}
\hline
\textbf{Parameter}& 
\textbf{Value} \\
\hline
Number of generations& 
51 \\
%\hline
Crossover probability& 
0.9 \\
%\hline
Mutations& 
2 / chromosome \\
%\hline
Function set& 
$F$ = \{+, -, *, /, \} \\
%\hline
Terminal set& 
Problem inputs + 4 supplementary registers  \\
%\hline
Selection& 
Binary Tournament \\
\hline
\end{tabular}
\end{center}
\end{table}

For all problems the relationship between the success rate, the 
chromosome length and the population size is analyzed. The success rate is 
computed as the number of successful runs over the total number of runs as described in section \ref{sec:15-2}. \\

\textbf{Experiment 1}\\

\begin{figure}[htbp]
\centerline{\includegraphics[width = \textwidth]{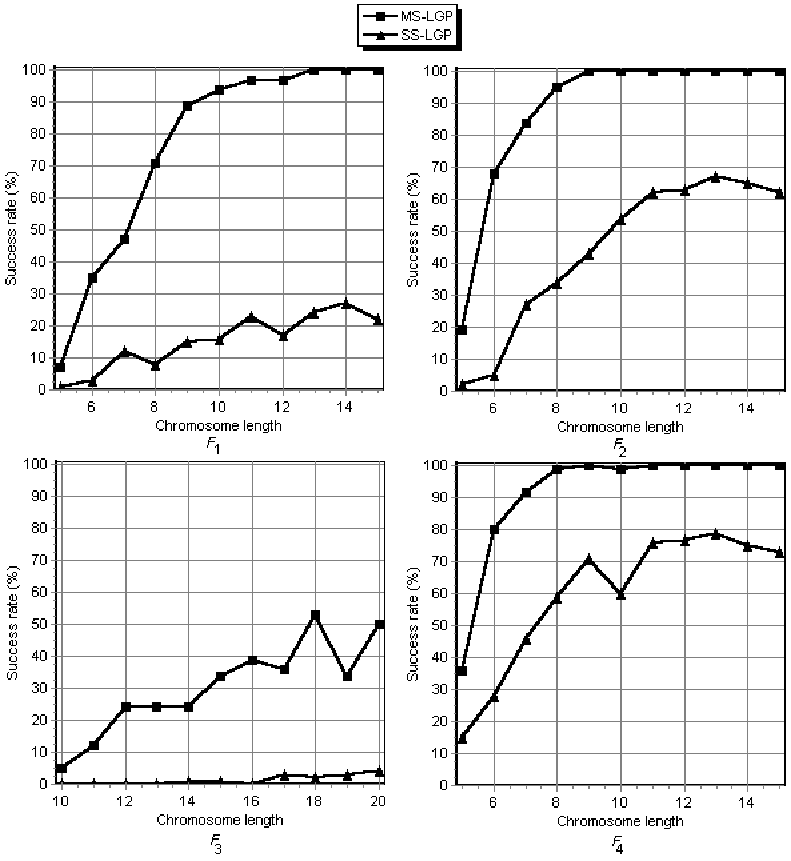}}
\caption{The relationship between the success rate and the number 
of instructions in a chromosome. Results are averaged over 100 runs.}
\label{fig:15-3}
\end{figure}

In this experiment the relationship between the success rate and the 
chromosome length is analyzed. The population size was 
set to 50 individuals. Other parameters of the LGP are given in 
Table \ref{tabel:15-6}. Results are depicted in Figure \ref{fig:15-3}.

Figure \ref{fig:15-3} shows that Multi-Solution LGP significantly outperforms 
Single-Solution LGP for all the considered test problems and for all the 
considered parameter setting. As in the case of MEP larger chromosomes are better 
for MS-LGP than shorter ones. This is due to the multi-solution 
ability: increasing the chromosome length leads to more solutions encoded in 
the same individual. The most difficult problem is $f_{3}$. For this problem the success rate of SS-LGP never increases over 5{\%}).\\

\textbf{Experiment 2}\\

\begin{figure}[htbp]
\centerline{\includegraphics[width = \textwidth]{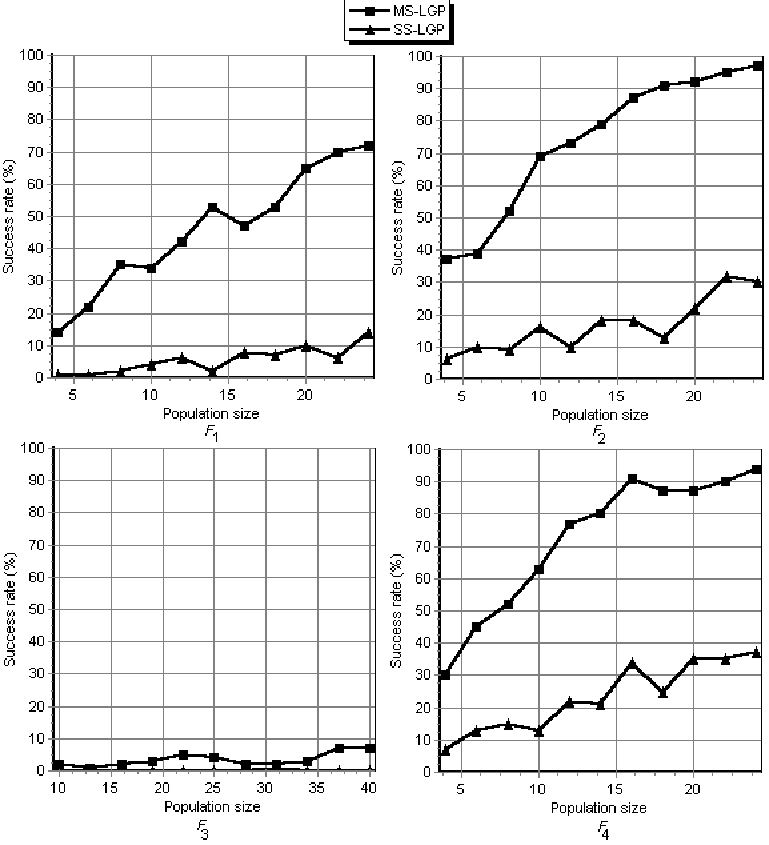}}
\caption{The relationship between the success rate and the population size. Results are averaged over 100 runs.}
\label{fig:15-4}
\end{figure}

In this experiment the relationship between the success rate and the 
population size is analyzed. The number of instructions 
in a LGP chromosome was set to 12. Other parameters for the LGP are given in Table \ref{tabel:15-6}. Results are depicted in Figure \ref{fig:15-4}.

Figure \ref{fig:15-4} shows that Multi-Solution LGP performs better 
than Single-Solution LGP. For the test problem $f_{3}$ the success rate of SS-LGP is 0{\%} (no run successfull).

\section{Infix Form Genetic Programming}\label{sec:15-5}

In this section \textit{Infix Form Genetic Programming} (IFGP) \cite{15:oltean3} technique is described. IFGP uses 
linear chromosomes for solution representation (i.e. a chromosome is a string of genes). An IFGP chromosome usually encodes several solutions of a problem.

In \cite{15:oltean3} IFGP was used for solving several classification problems taken from \cite{15:prechelt1}. The conclusion was that IFGP performs similar and sometimes even better that Linear Genetic Programming and Neural Networks \cite{15:brameier1,15:prechelt1}.

\subsection{Prerequisite}

We denote by $F$ the set of function symbols (or operators) that may appear in 
a mathematical expression. $F$ usually contains the binary operators {\{}+, $ - 
$, *, /{\}}. By \textit{Number{\_}of{\_}Operators} we denote the number of elements in $F$. A correct 
mathematical expression also contains some terminal symbols. The set of 
terminal symbols is denoted by $T$. The number of terminal symbols is denoted 
by \textit{Number{\_}of{\_}Variables}.

Thus, the symbols that may appear in a mathematical expression are $T \quad  \cup $ 
$F \quad  \cup $ {\{}'(', ')'{\}}. The total number of symbols that may appear in a 
valid mathematical expression is denoted by \textit{Number{\_}of{\_}Symbols}.

By $C_{i}$ we denote the value on the $i^{th}$ gene in a IFGP chromosome and 
by $G_{i}$ the symbol in the $i^{th}$ position in the mathematical expression 
encoded into an IFGP chromosome.

\subsection{IFGP Individual Representation}\label{sec:15-6}

In this section we describe how IFGP individuals are represented and how 
they are decoded in order to obtain a valid mathematical expression.

Each IFGP individual is a fixed size string of genes. Each gene is an 
integer number in the interval [0 .. \textit{Number{\_}Of{\_}Symbols }- 1]. An IFGP individual can be 
transformed into a functional mathematical expression by replacing each gene 
with an effective symbol (a variable, an operator or a parenthesis).\\

\textbf{Example}\\

If we use the set of functions symbols $F$ = {\{}+, *, -, /{\}}, and the 
set of terminals $T$ = {\{}$a$, $b${\}}, the following chromosome\\

C = 7, 3, 2, 2, 5\\

\noindent
is a valid chromosome in IFGP system.

\subsection{IFGP Decoding Process}

We begin to decode this chromosome into a valid mathematical 
expression. In the first position (in a valid mathematical expression) we 
may have either a variable, or an open parenthesis. That means that we have 
\textit{Number{\_}Of{\_}Variables} + 1 possibilities to choose a correct symbol on the first position. We put 
these possibilities in order: the first possibility is to choose the 
variable $x_{1}$, the second possibility is to choose the variable $x_{2}$ 
\ldots the last possibility is to choose the closed parenthesis ')'. The 
actual value is given by the value of the first gene of the chromosome. 
Because the number stored in a chromosome gene may be larger than the number 
of possible correct symbols for the first position we take only the value of 
the first gene \textit{modulo} number of possibilities for the first gene. Note that the \textit{modulo} operator is used in a similar context by Grammatical Evolution \cite{15:ryan1}. 

Generally, when we compute the symbol stored in the $i^{th}$ position in 
expression we have to compute first how many symbols may be placed in that 
position. The number of possible symbols that may be placed in the current 
position depends on the symbol placed in the previous position. Thus:

\begin{itemize}

\item[{\it (i)}]{if the previous position contains a variable ($x_{i})$, then for the current 
position we may have either an operator or a closed parenthesis. The closed 
parenthesis is considered only if the number of open parentheses so far is 
larger than the number of closed parentheses so far.}
\item[{\it (ii)}]{if the previous position contains an operator, then for the current position 
we may have either a variable or an open parenthesis. }
\item[{\it (iii)}]{if the previous position contains an open parenthesis, then for the current 
position we may have either a variable or another open parenthesis.}
\item[{\it (iv)}]{if the previous position contains a closed parenthesis, then for the current 
position we may have either an operator or another closed parenthesis. The 
closed parenthesis is considered only if the number of open parentheses so 
far is larger than the number of closed parentheses.}

\end{itemize}

Once we have computed the number of possibilities for the current position 
it is easy to determine the symbol that will be placed in that position: 
first we take the value of the corresponding gene modulo the number of 
possibilities for that position. Let $p$ be that value ($p$ = C$_{i}$ \textit{mod} 
\textit{Number{\_}Of{\_}Possibilities}). The $p^{th}$ symbol from the permitted symbols for the current is placed 
in the current position in the mathematical expression. (Symbols that may 
appear into a mathematical expression are ordered arbitrarily. For instance 
we may use the following order: $x_{1}$, $x_{2}$, \ldots , +, -, *, /, 
'(', ')'. )

All chromosome genes are translated excepting the last one. The last gene is used 
by the correction mechanism (see below).

The obtained expression usually is syntactically correct. However, in some 
situations the obtained expression needs to be repaired. There are two cases 
when the expression needs to be corrected:

The last symbol is an operator (+, -, *, /) or an open parenthesis. In 
that case a terminal symbol (a variable) is added to the end of the 
expression. The added symbol is given by the last gene of the chromosome.

The number of open parentheses is greater than the number of closed 
parentheses. In that case several closed parentheses are automatically added 
to the end in order to obtain a syntactically correct expression. \\

\textbf{Remark}. If the correction mechanism is not used, the last gene of the chromosome 
will not be used.\\

\textbf{Example}\\

Consider the chromosome $C$ = 7, 3, 2, 0, 5, 2 and the set of terminal and 
function symbols previously defined ($T$ = {\{}$a$, $b${\}}, $F$ = {\{}+, -, *, 
/{\}}).

For the first position we have 3 possible symbols ($a$, $b$ and '('). Thus, the 
symbol in the position $C_{0}$ \textbf{\textit{mod}} 3 = 1 in the array of 
possible symbols is placed in the current position in expression. The chosen 
symbol is b, because the index of the first symbol is considered to be 0.

For the second position we have 4 possibilities (+, -, *, /). The 
possibility of placing a closed parenthesis is ignored since the difference 
between the number of open parentheses and the number of closed parentheses 
is zero. Thus, the symbol '/' is placed in position 2.

For the third position we have 3 possibilities ($a$, $b$ and '('). The symbol 
placed on that position is an open parenthesis '('. 

In the fourth position we have 3 possibilities again ($a$, $b$ and '('). The 
symbol placed on that position is the variable $a$.

For the last position we have 5 possibilities (+, -, *, /) and the 
closed parenthesis ')'. We choose the symbol on the position 5 \textbf{mod} 
5 = 0 in the array of possible symbols. Thus the symbol '+' is placed in 
that position.

The obtained expression is $E=b$ / ($a$+.

It can be seen that the expression $E$ it is not syntactically correct. For 
repairing it we add a terminal symbol to the end and then we add a closed 
parenthesis. Now we have obtained a correct expression:

\begin{center}
$E=b/(a+a)$.
\end{center}

\begin{figure}[htbp]
\centerline{\includegraphics{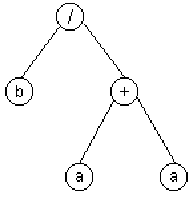}}
\caption{The expression tree of $E=b$ / ($a+a)$.}
\label{fig:15-5}
\end{figure}

The expression tree of $E$ is depicted in Figure \ref{fig:15-5}.

\subsection{Fitness Assignment Process}

In this section we describe how IFGP may be efficiently used for solving 
symbolic regression problems.

A GP chromosome usually stores a single solution of a problem and the 
fitness is normally computed using a set of fitness cases. 

Instead of encoding a single solution, an IFGP individual is allowed to 
store multiple solutions of a problem. The fitness of each solution is 
computed in a conventional manner and the solution having the best fitness 
is chosen to represent the chromosome. 

In the IFGP representation each sub-tree (sub-expression) is considered as a 
potential solution of a problem. For example, the previously obtained 
expression (see section \ref{sec:15-6}) contains 4 distinct solutions 
(sub-expressions):
\\

$E_{1}=a$,

$E_{2}=b$,

$E_{3}=a+a$,

$E_{4}=b / (a+a)$.
\\

The fitness of an IFGP individual using the same formulas used for MEP in section \ref{sec:15-3-2}.

\subsection{Search Operators}

Search operators used within the IFGP model are recombination and mutation. 
These operators are similar to the genetic operators used in conjunction 
with binary encoding \cite{15:dumitrescu1}. By recombination two parents exchange genetic 
material in order to obtain two offspring. In this paper only two-point 
recombination is used. Mutation operator is applied with a fixed 
mutation probability ($p_{m})$. By mutation a randomly generated value over 
the interval [0, \textit{Number{\_}of{\_}Symbols}-1] is assigned to the target gene.

\subsection{IFGP Algorithm}

A steady-state \cite{15:syswerda1} algorithm (similar to that used by MEP) is used as underlying mechanism for IFGP. The algorithm starts with a randomly chosen population of individuals. The following steps are repeated until a termination condition is reached. Two parents are chosen at each step using binary tournament selection \cite{15:dumitrescu1}. The selected individuals are recombined with a fixed crossover probability $p_{c}$. By 
recombining two parents, two offspring are obtained. The offspring are 
mutated and the best of them replaces the worst individual in the current 
population (only if the offspring is better than the worst individual in 
population). 

The algorithm returns as its answer the best expression evolved 
for a fixed number of generations.

\subsection{Single Solution Infix Form Genetic Programming}

As IFGP was originally designed to encode multiple solutions in a single chromosome \cite{15:oltean3} we modify the technique in order to encode a single solution / chromosome. The obtained variant is called Single-Solution IFGP (SS-IFGP). The expression encoded in a SS-IFGP chromosome is that representing the entire individual (the largest expression).\\

\textbf{Example}\\

Consider again the chromosome:\\

C = 7, 3, 2, 2, 5.\\

The SS-IFGP expression is:\\

$E=b / (a+a)$.

\subsection{Numerical Experiments with IFGP and SS-IFGP}

Several numerical experiments with Infix Form Genetic Programming and Single-Solution Infix Form Genetic Programming are performed in this section using the test problems described in section \ref{sec:15-2}.

The general parameters of the IFGP and SS-IFGP algorithms are given in Table \ref{tabel:15-7}. The same 
settings are used for Infix Form Genetic Programming and for Single Solution Infix Form Genetic Programming.

\begin{table}[htbp]
\begin{center}
\caption{Parameters of the IFGP and SS-IFGP algorithms for solving symbolic regression problems.}
\label{tabel:15-7}
\begin{tabular}
{p{130pt}p{130pt}}
\hline
\textbf{Parameter}& 
\textbf{Value} \\
\hline
Number of generations& 
51 \\
%\hline
Crossover probability& 
0.9 \\
%\hline
Mutations& 
2 / chromosome \\
%\hline
Function set& 
$F = \{+, -, *, /\}$ \\
%\hline
Terminal set& 
Problem inputs\\
%\hline
Selection& 
Binary Tournament \\
\hline
\end{tabular}
\end{center}
\end{table}

For all problems the relationship between the success rate and the 
chromosome length and the population size is analyzed. The success rate is 
computed as the number of successful runs over the total number of runs. \\

\textbf{Experiment 1}\\

\begin{figure}[htbp]
\centerline{\includegraphics[width = \textwidth]{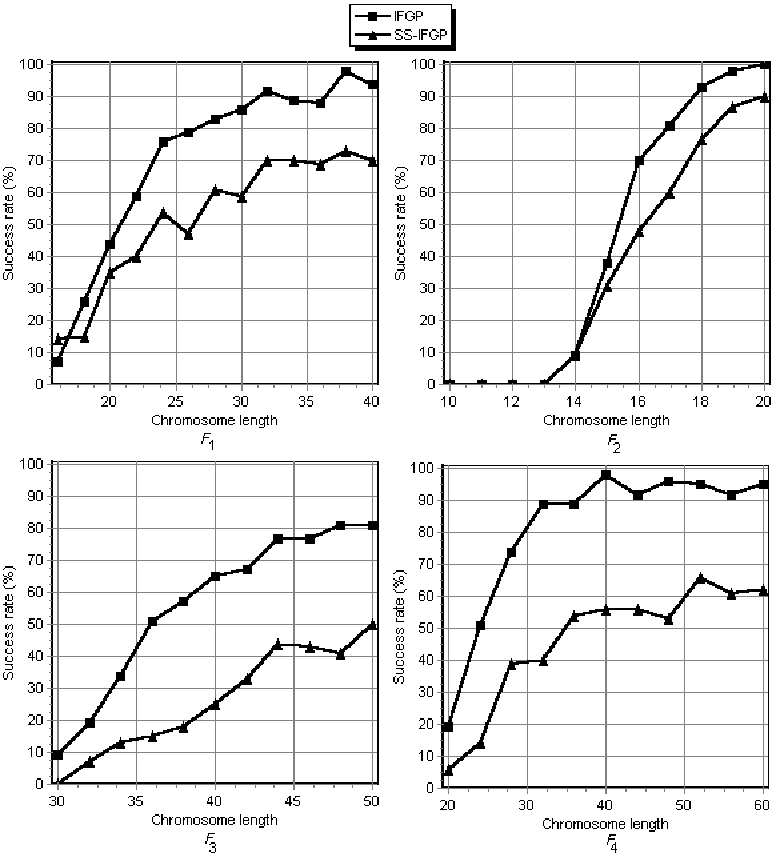}}
\caption{The relationship between the success rate and the number 
of symbols in a chromosome. Results are averaged over 100 runs.}
\label{fig:15-6}
\end{figure}

In this experiment the relationship between the success rate and the 
chromosome length is analyzed. The population size was 
set to 50 individuals. Other parameters of the IFGP and SS-IFGP algorithms are given in 
Table \ref{tabel:15-7}. Results are depicted in Figure \ref{fig:15-6}.

Figure \ref{fig:15-6} shows that Infix Form Genetic Programming outperforms 
Single-Solution Infix Form Genetic Programming for the considered test problems. However the differences between MS-IFGP and SS-IFGP are not so significant as in the case of MEP and LGP.\\

\textbf{Experiment 2}\\

\begin{figure}[htbp]
\centerline{\includegraphics[width = \textwidth]{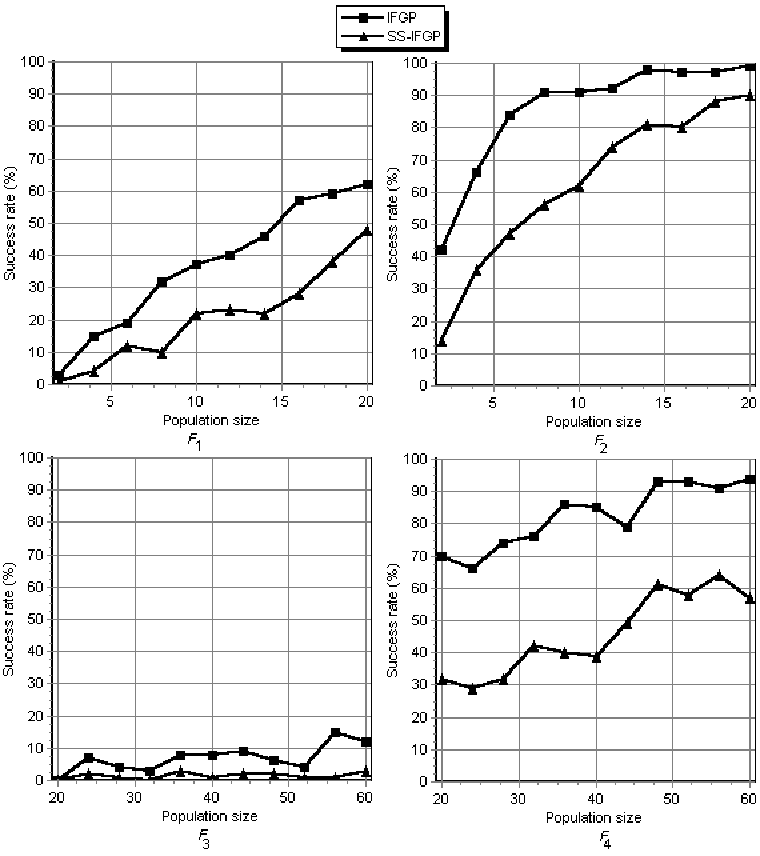}}
\caption{The relationship between the success rate and the population size. Results are averaged over 100 runs.}
\label{fig:15-7}
\end{figure}

In this experiment the relationship between the success rate and the 
population size is analyzed. The number of symbols in a IFGP or SS-IFGP chromosome was set to 30. Other parameters for the IFGP and SS-IFGP are given in Table \ref{tabel:15-7}. Results are depicted in Figure \ref{fig:15-7}.

Figure \ref{fig:15-7} shows that Infix Form Genetic Programming performs better 
than Single-Solution Infix Form Genetic Programming. Problem $f_{2}$ is the easiest one (the IFGP success is 100 {\%} when the population size is 20) when the chromosome length is 20 and problem $f_{3}$ is the most difficult one.

\section{Conclusions and Further Work}
\label{sec:15-6}

The ability of encoding multiple solutions in a single chromosome has been analyzed in this paper for 3 GP techniques: Multi Expression Programming, Linear Genetic Programming and Infix Form Genetic Programming. It has been show how to efficiently decode the considered chromosomes by traversing them only once. 

Numerical experiments have shown that Multi-Solution Programming significantly improves the evolutionary search for all the considered test problems. There are several reasons for which Multi Solution Programming performs better than Single Solution Programming:

\begin{itemize}

\item{MSP chromosomes act like variable-length chromosomes even if they are stored as fixed-length chromosomes. The variable-length chromosomes are better than fixed-length chromosomes because they can easily store expressions of various complexities,}

\item{MSP algorithms perform more function evaluations than their SSP counterparts. However the complexity of decoding individuals is the same for both MSP and SSP techniques.}

\end{itemize}

The multi-solution ability will be investigated within other evolutionary models.

\end{document}